\documentclass{article}
\usepackage{ijcai26}

\usepackage{times}
\usepackage{soul}
\usepackage{url}
\usepackage[hidelinks]{hyperref}
\usepackage[utf8]{inputenc}
\usepackage[small]{caption}
\usepackage{graphicx}
\usepackage{amsmath}
\usepackage{amsthm}
\usepackage{booktabs}
\usepackage{algorithm}
\usepackage{algorithmic}
\usepackage[switch]{lineno}
\usepackage{xcolor}
\usepackage{bm}

\usepackage[table]{xcolor}  

\usepackage{array} 
\usepackage{tabularx} 
\usepackage{amssymb}

\usepackage{xspace}

\linenumbers

\title{Powerful Teachers Matter: Text-Guided Multi-view Knowledge Distillation with Visual Prior Enhancement}

\author{
Xin Zhang$^1$\and
Jianyang Xu$^1$\and
Hao Peng$^{1}$\and
Dongjing Wang$^1$\and
Jingyuan Zheng$^1$\and
Yu Li$^1$\and
Yuyu Yin$^1$\and
Hongbo Wang$^1$
\affiliations
$^1$Hangzhou Dianzi University\\
\emails
\{zhangxin,xujianyang,penghao,dongjing.wang,zhengjoy,liyucomp,yinyuyu,whongbo,chentao11\}@hdu.edu.cn
}

\begin{document}
\nolinenumbers

\maketitle

\newcommand{\proposedmethod}{TMKD\xspace}

\begin{abstract}

Knowledge distillation transfers knowledge from large teacher models to smaller students for efficient inference. While existing methods primarily focus on distillation strategies, they often overlook the importance of enhancing teacher knowledge quality. In this paper, we propose Text-guided Multi-view Knowledge Distillation (\proposedmethod), which leverages dual-modality teachers, a visual teacher and a text teacher (CLIP), to provide richer supervisory signals. Specifically, we enhance the visual teacher with multi-view inputs incorporating visual priors (edge and high-frequency features), while the text teacher generates semantic weights through prior-aware prompts to guide adaptive feature fusion. Additionally, we introduce vision-language contrastive regularization to strengthen semantic knowledge in the student model. Extensive experiments on five benchmarks demonstrate that \proposedmethod consistently improves knowledge distillation performance by up to 4.49\%, validating the effectiveness of our dual-teacher multi-view enhancement strategy. Code is available at \url{https://anonymous.4open.science/r/TMKD-main-44D1/}.

\end{abstract}

\section{Introduction}
\begin{figure}[ht!]
    \centering
    \includegraphics[width=1\linewidth]{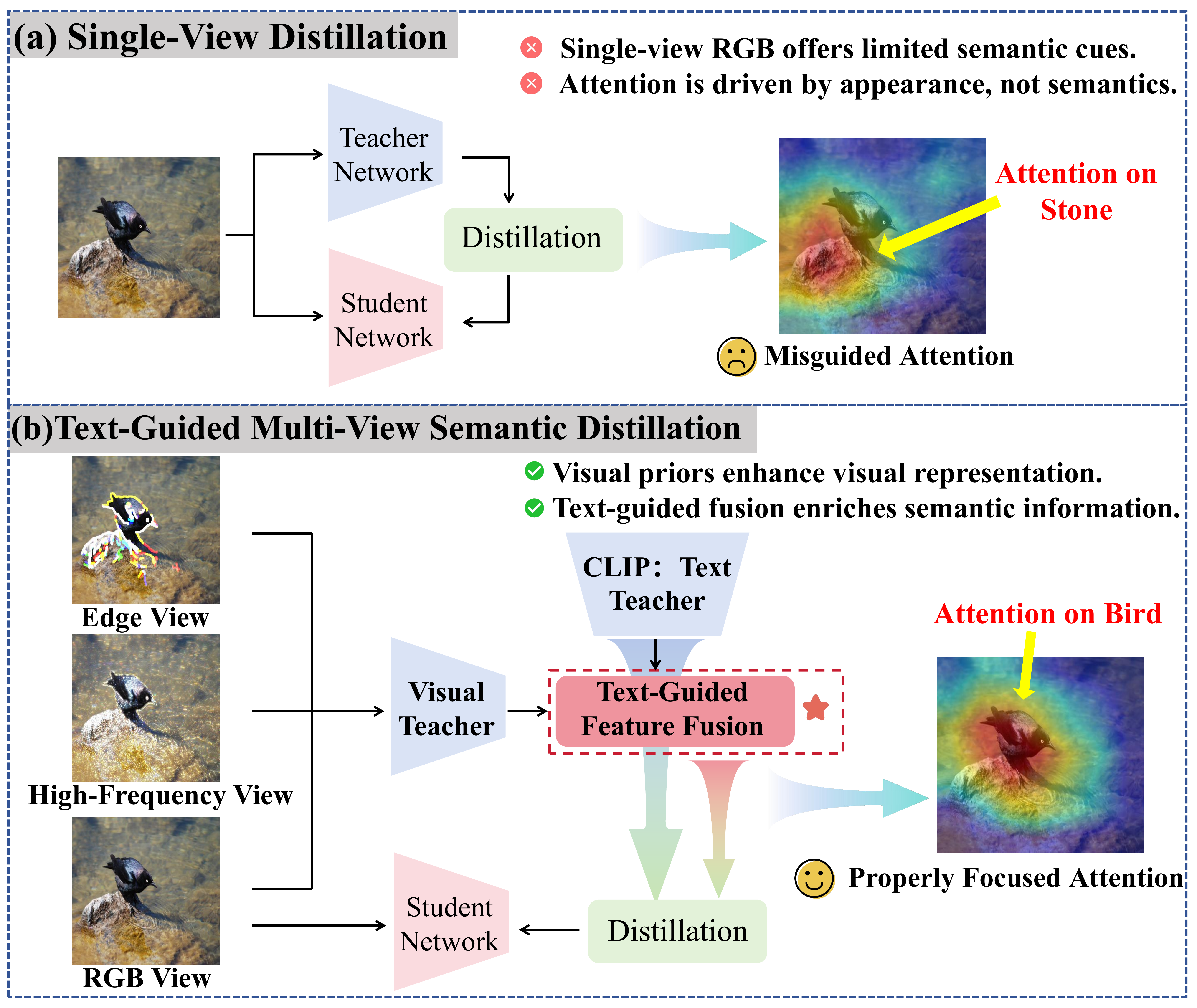}
    \caption{
(a) Single-view Distillation: Attention incorrectly focuses on the stone rather than the bird.
(b) Our Multi-view Distillation: Attention correctly targets the bird, demonstrating that the model captures task-relevant features.
    }

    \label{fig:kd_comparison}
\end{figure}
Knowledge distillation (KD) \cite{hinton2015distilling} is a key technology for model compression \cite{gou2021knowledge}, enabling lightweight student models to learn from high-capacity teacher models. It has achieved notable success in computer vision tasks such as image classification \cite{SunWHC2025,BaiLKYY2026,Zhang2024selfkd}, object detection \cite{Cai2026ODKD}, and semantic segmentation \cite{Zi2023catkd}. However, most existing methods rely on single-view teacher models as shown in Fig.\ref{fig:kd_comparison}(a), which limits their ability to fully exploit the rich semantic information present in images and misdirect attention toward irrelevant background regions (e.g., rocks). As a result, the knowledge distilled to the student model is often suboptimal, restricting the potential of knowledge distillation.

In real-world scenarios, images contain rich \cite{Guo2023Speakerextract} and complementary visual cues \cite{Kong2025CLIPExtract}, making it difficult to capture the full scene content only using RGB image. To address this, multi-view representations \cite{Wang2025SeqMvRL,Bao2024MVF} are generated through various visual preprocessing methods, such as edge-enhanced views \cite{Yang2025edge}, and high-frequency detail enhanced views \cite{Cantone2026sal_guide}, each highlighting different aspects of visual information. However, the absence of a unified semantic alignment mechanism  remains a challenge. Inconsistency between different views can lead to conflicting supervision signals and hinder effective knowledge transfer.

In recent years, vision-language models \cite{Chen2025LRME,Chen2025VLM} such as CLIP \cite{CLIP} have demonstrated strong capabilities in aligning heterogeneous \cite{Li2025Tackling,Hong2025Aggregation} visual modalities by providing interpretable and modality-independent semantic priors in a shared visual-text embedding space \cite{Wei2023ELITE}. However, existing CLIP-based KD methods typically use CLIP as a general teacher or as a category-level supervision signal \cite{Jang2025VL2Lite}. 
Its potential to guide modality-aware feature fusion through semantic alignment remains largely unexplored.

\begin{figure*}[ht!]
    \centering
    \includegraphics[width=1\linewidth]{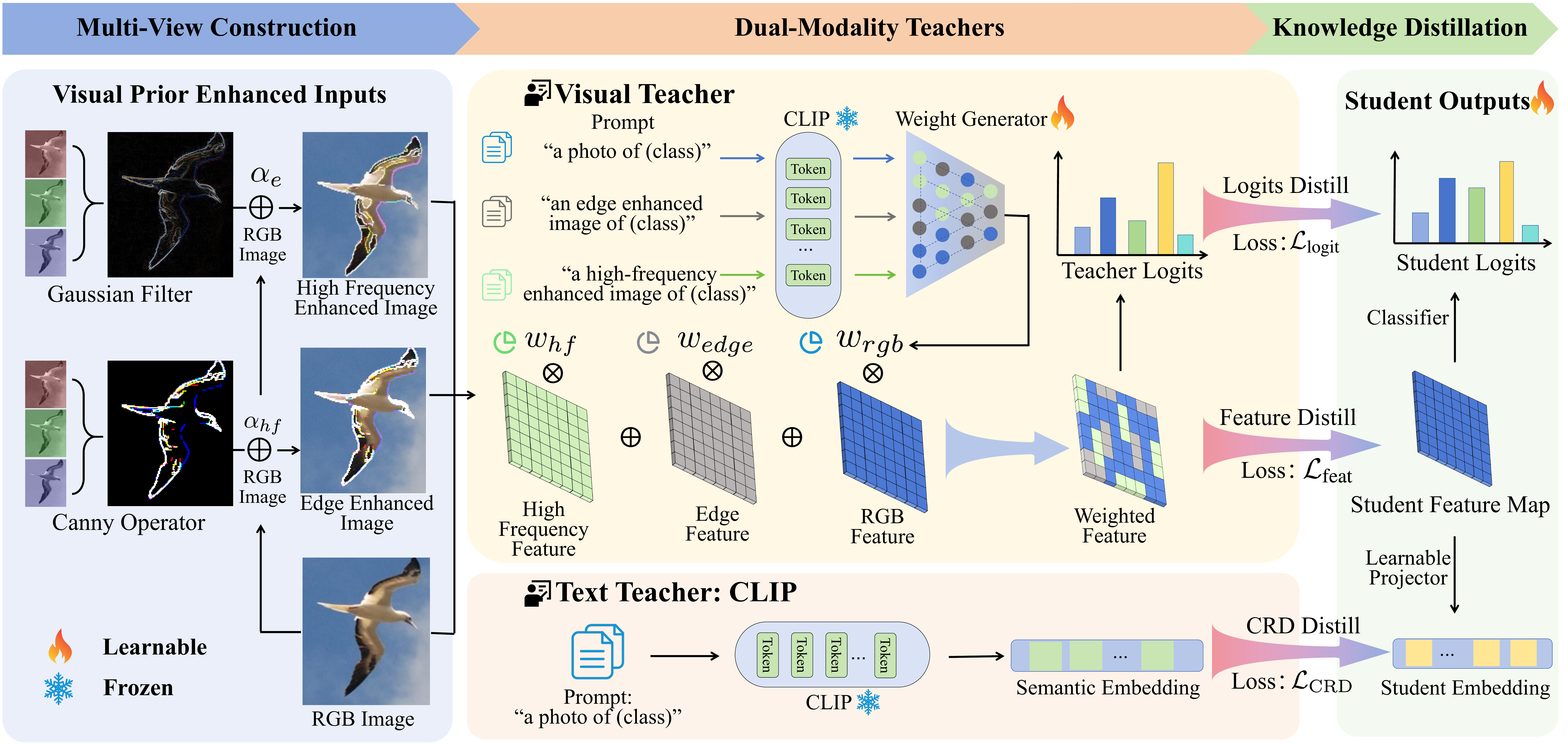}
    \caption{Overview of our \proposedmethod framework. \proposedmethod employs dual-modality teachers: (1) a visual teacher that extracts features from multi-view inputs and  fuses them with semantic weights through prior-aware prompts, and (2) a text teacher (CLIP).  
    The visual teacher transfers knowledge through feature-level and logits-level distillation, while the text teacher provides semantic guidance via vision-language contrastive representation distillation (CRD) loss, enabling effective transfer of both visual and semantic knowledge to the student model.}
    \label{fig:architercture_map}
\end{figure*}

To address the above issues, we propose a text-guided multi-view knowledge distillation framework (\proposedmethod) (see Fig.~\ref{fig:architercture_map}), which leverages dual-modality teachers, a multi-view visual teacher and a text teacher (CLIP), to provide richer supervisory signals. Starting from a single RGB image, we generate two complementary views: an edge-enhanced view capturing structural information and a high-frequency view highlighting fine-grained details. The visual teacher extracts rich features from these three views (RGB, edge, and high-frequency). Meanwhile, we use CLIP to generate semantic weights through prior-aware prompts, enabling adaptive fusion of multi-view features based on their semantic relevance. Furthermore, we introduce vision-language contrastive regularization that aligns the student's visual representations with textual embeddings from the text teacher, strengthening semantic knowledge during distillation. By integrating visual priors with semantic guidance from dual-modality teachers, \proposedmethod produces more discriminative and semantically consistent features, facilitating effective knowledge transfer to the student model. As shown in Fig.~\ref{fig:kd_comparison}(b), \proposedmethod effectively focuses attention on the target object.

Our main contributions are summarized as follows:
\begin{itemize}
    \item We propose TMKD, a text-guided multi-view knowledge distillation framework. We leverage dual-modality teachers to enhance knowledge transfer by integrating complementary visual priors with CLIP-based semantic guidance, enabling more discriminative and semantically consistent feature learning.

    \item We introduce vision-language contrastive regularization that aligns student representations with textual embeddings, strengthening semantic knowledge acquisition.

    \item Extensive experiments on five datasets demonstrate that TMKD consistently improves state-of-the-art KD methods by up to 4.49\%, with comprehensive ablations and visualization validating the effectiveness.

\end{itemize}

\section{Related Work}

\subsection{Knowledge Distillation}
Knowledge distillation is a key model compression technique \cite{Mora2024FLKD,Zhao2025BRSKD}, transferring knowledge from a large teacher model to a smaller student model. Early methods focused on output-based distillation via softened logits, allowing the student to learn “dark knowledge” from the teacher. Subsequent research explored feature-based \cite{Zi2023catkd} and relation-based distillation \cite{Tung2019SPKD}, preserving intermediate representations or structural relationships between samples \cite{park2019relational,Tung2019SPKD}. Attention mechanisms further improved knowledge transfer by focusing on important feature maps.

Despite progress, most methods use only single-modality \cite{NiZ23IJCAIKD,Lee2025cvprkd} image representations, assuming all discriminative information is in the visual stream. In reality, images contain rich semantic cues, such as edges or high-frequency details, which are underutilized in traditional distillation. Existing methods fall into response-based \cite{Zhao2022DKD}, feature-based \cite{Zi2023catkd}, and relation-based \cite{Tung2019SPKD} categories, but often neglect these complementary cues.

\subsection{VLMs for Knowledge Transfer}
Vision language models like CLIP learn joint image-text embeddings \cite{Wu2025MRSKD}, enabling strong cross-modal alignment \cite{Eutamene2024SIDVLM,Li2022ISSKD} and robust zero-shot performance \cite{Liu2024LLMKD}. CLIP has been widely used for downstream tasks, including knowledge distillation \cite{Ta2025MSCLIPKD,Zhou2025CLIPIKD}. CLIP-based distillation approaches mainly: (1) use CLIP as a teacher to transfer cross-modal knowledge \cite{Jang2025VL2Lite}, or (2) employ CLIP’s text supervision to align visual features.

Our approach differs by using CLIP as a semantic prior to guide multi-modal feature fusion within the teacher model. We design text prompts for different visual modalities and leverage CLIP’s embeddings to compute adaptive fusion weights, introducing semantic awareness without altering the teacher or loss function.

\subsection{Multi-Teacher Distillation}
Multi-teacher distillation methods integrate features from several teachers \cite{Liu2025NNMTKD,Ma2024CAGMKD,Jiang2024MTKD}, providing richer supervision but increasing complexity, computation, and possible conflicts \cite{Yang2025MTKD,Zhang2023MMKD}. In contrast, our method extracts multiple complementary feature modalities (RGB, edges, high-frequency details) from a single teacher. These modalities are semantically distinct and fused adaptively using CLIP’s text prior, enabling effective multi-modal supervision without extra teachers or complex aggregation, thus maintaining simplicity and efficiency.

\section{Method}
\label{sec3}
We propose \proposedmethod, a knowledge distillation framework that employs dual-modality teachers, a visual teacher and a text teacher (CLIP), to provide enriched supervisory signals. Given a single RGB image, we generate multi-view inputs (RGB, edge, and high-frequency views) to capture complementary visual priors. The visual teacher extracts features from these views, which are adaptively fused using semantic weights generated by CLIP text encoder through prior-aware prompts. The visual teacher transfers knowledge through feature-level and logits-level distillation, while the text teacher CLIP enhances semantic understanding via vision-language contrastive distillation, enabling the student to learn rich representations. Fig.~\ref{fig:architercture_map} illustrates the overall framework.

\subsection{Multi-view Images Construction}

In order to fully exploit the complementary visual cues, we construct multi-view inputs from a single RGB image. They provide diversified supervision signals for the teacher network by emphasizing structure information and high-frequency regions respectively. 

\subsubsection{Edge Enhanced View Generation}
To enhance the structural cues, we construct edge enhanced view by edge detection and image-edge fusion. Given an RGB image $I \in \mathbb{R}^{H \times W \times 3}$, we first perform edge extraction using Canny operator for each color channel. The threshold is set to $(\theta_1, \theta_2) = (100, 200)$. The edge results of the three channels are concatenated to obtain a three-channel edge map $E$. Then, we get edge enhanced view by weighted superposition:
\begin{equation}
I_{\text{edge}} = \text{clip}\Big( \frac{I}{255} + \alpha_e \frac{E}{255}, 0, 1 \Big),
\end{equation}
where $\alpha_e = 1.5$ is used to adjust the weight of the edge result. $\text{clip}(\cdot)$ is used to ensure that the fused pixel values are within the legal range. The fusion strategy effectively strengthens the edge and contour information in the image while preserving the original color distribution.

\subsubsection{High-Frequency Enhanced View Generation}
High-frequency information plays a crucial role in the semantic understanding of images. Therefore, we construct a high-frequency view input to enhance the representation capability of the teacher model. Specifically, given an RGB image $I \in \mathbb{R}^{H \times W \times 3}$, Gaussian filtering is first applied to each color channel separately to obtain a smoothed version. The residual between the original channel and its smoothed result is treated as the high-frequency response $S$ of the corresponding channel. Then, we perform weighted fusion:
\begin{equation}
I_{\text{hf}} = \text{clip}\Big( \frac{I}{255} + \alpha_{hf} \frac{S}{255}, 0, 1 \Big),
\end{equation}
where $\alpha_{hf} = 1.5$ is used to control the enhancement of high-frequency information. $I_{\text{hf}}$ significantly magnifies texture details and fine-grained structural information, thus forming an effective complementary view for RGB image and edge-enhanced view image.

\subsection{Text-guided Feature Fusion}
\label{subsec:semantic_fusion}

We extract the feature representations of each view by one shared teacher network, which are denoted as $F_{\text{RGB}}$, $F_{\text{edge}}$ and $F_{\text{hf}}$, respectively. These features are derived from complementary visual priors and encode discriminative cues related to semantic appearance, geometric structure and attention respectively. Then, the above multi-view features are fused under the semantic guidance condition to generate a unified teacher representation $F_{\text{fused}}$, which is used as the core signal to supervise the student model.

\subsubsection{View-specific Prompts Construction} To incorporate explicit semantic priors into multi-view feature fusion, we design view-specific prompts that capture the distinct visual characteristics of each input view. Rather than relying on generic category labels, we construct specialized prompts that provide CLIP's text encoder with view-aware semantic context, enabling more precise alignment between textual and visual representations.

Specifically, for each input image, we generate three tailored prompts corresponding to its class label, as detailed in Table \ref{tab:prompt_examples}. Each prompt emphasizes the unique properties of its corresponding view. RGB view captures appearance and color information. The edge view focuses on structural boundaries and contours, while the high-frequency view highlights texture details and fine-grained patterns. This view-specific prompt design ensures that CLIP's text encoder produces discriminative embeddings that align with the semantic characteristics of each visual modality.

\begin{table}[t]
\scriptsize
\setlength{\tabcolsep}{6pt}
\renewcommand{\arraystretch}{1.1}
\centering
\begin{tabularx}{\columnwidth}{l X}
\toprule
\textbf{Input View} & \textbf{Prompt} \\
\midrule
RGB & \textit{``a photo of a \{class\}''} \\
Edge View & \textit{``an edge enhanced image of a \{class\}''} \\
High-Frequency View & \textit{``a high-frequency enhanced image of a \{class\}''} \\
\bottomrule
\end{tabularx}
\caption{View-specific prompts constructed for CLIP guidance.}
\label{tab:prompt_examples}
\end{table}

\subsubsection{Semantic-Guided Adaptive Fusion}
View-specific text prompts are encoded into semantic embeddings $t_{\mathrm{rgb}}$, $t_{\mathrm{edge}}$, and $t_{\mathrm{hf}}$ using CLIP's text encoder. Leveraging CLIP's pre-training on large-scale vision-language data, these embeddings maintain semantic alignment across views while preserving their distinct visual characteristics, providing reliable semantic guidance for multi-view feature fusion.

The three text embeddings are processed by a lightweight weight generator, a two-layer fully connected network, to produce normalized fusion weights that dynamically adjust each view's contribution:
\begin{equation}
    x = \mathrm{ReLU}\Big(W_1 [\, t_{\mathrm{rgb}}; t_{\mathrm{edge}}; t_{\mathrm{hf}} \,] + b_1 \Big),
\end{equation}
\begin{equation}
    w = \mathrm{Softmax}(W_2 x + b_2),
\end{equation}
where $w=[w_{rgb},w_{edge},w_{hf}]\in \mathbb{R}^3$ denotes the normalized fusion weights for RGB, edge-enhanced, and high-frequency views respectively. $W_1$, $W_2$ and $b_1$, $b_2$ are the weights and biases of the fully connected layers.

The multi-view teacher features $F_{\mathrm{rgb}}$, $F_{\mathrm{edge}}$, and $F_{\mathrm{hf}}$ are then adaptively fused using these semantic weights:
\begin{equation}
    F_{\mathrm{fused}} = w_{\mathrm{rgb}} F_{\mathrm{rgb}} + w_{\mathrm{edge}} F_{\mathrm{edge}} + w_{\mathrm{hf}} F_{\mathrm{hf}}.
\end{equation}

This adaptive fusion mechanism dynamically emphasizes information-rich views while preserving complementary structural and semantic features from all image views. The resulting $F_{\mathrm{fused}}$ serves as the primary supervision signal for feature-level distillation. It enables the student to learn both semantic understanding and multi-view complementary information, including structural patterns and high-frequency details, thereby achieving more robust and discriminative representations.

\subsection{Distillation Loss Function}
We employ three complementary distillation strategies: feature-level distillation to transfer the teacher's internal representations, logits-level distillation to align prediction distributions, and text-guided contrastive distillation to anchor student features to CLIP's semantic space.

\subsubsection{Feature-level Distillation}
To enable the student model to stably learn semantic patterns encoded by the teacher across different visual views $F_{\mathrm{fused}}$, we introduce small random perturbations to enhance robustness and prevent overfitting:

\begin{equation}
    \tilde{F}_{\text{fused}} = F_{\text{fused}} + \gamma \epsilon,
\end{equation}
where $\epsilon \sim \mathcal{N}(0, I)$ denotes Gaussian noise and $\gamma$ is a scaling factor.

Instead of directly using the $\ell_2$ loss to regression the feature vectors, we adopt the temperature scaled Kullback-Leibler (KL)  divergence to align the student feature distribution with the perturbed teacher feature distribution. Let $F_s$ be the feature representation of the student model after learnable mapping, and $\tau_{f}$ be the distillation temperature. The feature distillation loss function is defined as follows:
\begin{equation}
    \mathcal{L}_{\text{feat}}
    = \tau_{f}^2 \, \mathrm{KL} \left(
        \log\mathrm{Softmax}\left(\frac{F_s}{\tau_{f}}\right)
        \,\middle\|\, 
        \mathrm{Softmax}\left(\frac{\tilde{F}_{\text{fused}}}{\tau_{f}}\right)
    \right)
\end{equation}

This design offers two key advantages. First, Gaussian noise regularization improves the student's generalization to unseen inputs and enhances robustness against teacher prediction errors. Second, KL divergence enables distribution-level matching, allowing the student to capture fine-grained semantic variations rather than simply imitating feature magnitudes. These mechanisms jointly ensure effective knowledge transfer of complementary structural, textural, and semantic information from the multi-view teacher, yielding richer and more discriminative representations.

\subsubsection{Logits-level Distillation}
We apply standard logits distillation to align the class prediction distributions between student and teacher models. The logits distillation loss is defined using KL divergence:
\begin{equation}
\mathcal{L}_{\text{logit}} = 
\mathrm{KL}\big(
\mathrm{Softmax}(\mathbf{z}_t / \tau_l)
\;\big\|\;
\mathrm{Softmax}(\mathbf{z}_s / \tau_l)
\big),
\end{equation}
where $\mathbf{z}_t$ and $\mathbf{z}_s$ denote the teacher and student logits respectively, and $\tau_l$ is the distillation temperature.

This objective enables the student to inherit the teacher's category-level discriminative knowledge, including inter-class similarity structures, while providing global semantic supervision that complements feature-level distillation.

\subsubsection{Text-guided Contrastive Representation Distillation}
In addition to the supervision of the feature and output layers, we introduce a text-guided contrastive representation distillation (CRD)  objective for distilling semantic knowledge from the text teacher CLIP. Different from traditional CRD  \cite{Yong2019CRD} method that only compare between student and teacher features, our method uses CLIP text embeddings as semantic anchors to guide students to learn representations that are both discriminative and semantically robust.

Specifically, given a student feature $F_s$, we map it to a low-dimensional embedding space by a learnable projection head $g(\cdot)$ and perform $\ell_2$ normalization:
\begin{equation}
z_s = g(F_s) / \|g(F_s)\|_2.
\end{equation}

Meanwhile, for each training sample, its corresponding class-aware prompt \textit{``a photo of a \{class\}''} is encoded by the CLIP text encoder to obtain the semantic text embedding $t$, which is also projected and normalized:
\begin{equation}
z_{\text{text}} = g(t) / \|g(t)\|_2.
\end{equation}

Subsequently, we employ a contrastive learning loss term to guide student embeddings $z_s^i$ close to their matched text embeddings $z_{\mathrm{text}}^i$ while moving away from the unmatched text embeddings stored in the memory bank. For mini-batches of size $B$, the CRD loss is defined as follows:

\begin{equation}
\begin{aligned}
\mathcal{L}_{\mathrm{CRD}} 
= - \frac{1}{B} \sum_{i=1}^{B} \Big[ 
& \log \frac{\exp(s_{ii}/\tau_{crd})}{\sum_{j=1}^{B} \exp(s_{ij}/\tau_{crd})} \\
& + \log \frac{\exp(s_{ii}/\tau_{crd})}{\sum_{j=1}^{B} \exp(s_{ji}/\tau_{crd})}
\Big],
\end{aligned}
\end{equation}
where $s_{ij} = z_s^i \cdot z_{\text{text}}^j$, $s_{ji} = z_{\text{text}}^i \cdot z_s^j$, and $\tau_{crd}$ is the temperature parameter. Negative samples are sampled from the CLIP text embedding memory.
Through this text-guided contrast objective, student visual representations are explicitly anchored to the CLIP semantic space at the instance level, enabling semantic alignment. This CRD objective not only complements the supervision signal distilled by the teacher feature layer, but also makes the student representation more semantically interpretable and robust while maintaining visual discrimination consistency.

\subsubsection{Overall Loss Function}
The overall training loss function of our method consists of a weighted combination of feature-level distillation, logits-level distillation, and text-guided contrastive distillation:


\begin{equation}
    \mathcal{L}_{\text{all}} 
    = \alpha \, \mathcal{L}_{\mathrm{logit}}
    + \beta \, \mathcal{L}_{\mathrm{CRD}}
    + \gamma \, \mathcal{L}_{\mathrm{feat}} .
\end{equation}

The coefficients $\alpha$, $\beta$ and $\gamma$ are used to adjust the relative weights of each loss term, thereby balancing the contributions of the three types of supervision signals (features, logits and semantic contrast) in the training process.

\section{Experiments}
We evaluate our \proposedmethod method through: (1) comparisons with state-of-the-art KD methods on multiple benchmarks, (2) generalization tests across diverse architectures, (3) ablation studies on key components, and (4) visualization analysis. Results demonstrate that our multi-view semantic enhancement consistently improves distillation performance.

\subsection{Experimental Setup}
\subsubsection{Datasets}
We conduct experiments on five datasets, covering both fine-grained and general image classification:
CUB-200 (bird species), RAF-DB (facial expressions), CIFAR-100 (general object classification), DTD (textures) and Stanford Dogs (dog breeds). 

\subsubsection{Baselines}
We compare our approach with representative knowledge distillation baselines:

\begin{itemize}
    \item KD \cite{hinton2015distilling}: Classic response-based distillation via softened logits.

    \item RKD \cite{park2019relational}: Transfers structural knowledge by modeling sample relations.

    \item SP \cite{Tung2019SPKD}: Preserves similarity structure of teacher features.

    \item CAT-KD \cite{Zi2023catkd}: Uses channel attention for enhanced feature transfer.

    \item TeKAP \cite{Md2025TeKAP}: Simulates diverse knowledge at the logits level without extra teachers.
\end{itemize}

    
    


\subsubsection{Implementation Details}
All experiments use multi-view inputs (RGB, edge, high-frequency maps) and support any teacher-student combination. For each sample, view-specific text prompts are encoded by CLIP (ViT-B/32) to provide semantic priors. Distillation uses fixed hyperparameters: temperature $\tau_l=4$, feature temperature $\tau_f=2$, contrastive representation distillation temperature $\tau_{crd}=2$, and loss weights $\alpha=2$, $\beta=0.01$, $\gamma=0.1$; CAT-KD has a weighting of 5. Feature alignment employs learnable projectors, and view fusion is guided by WeightNet using CLIP embeddings. Models are trained jointly with SGD (learning rate 0.001, momentum 0.9, weight decay $5 \times 10^{-4}$) for 120 epochs, using warmup and stepwise decay. The best student is selected by validation accuracy.

\subsection{Comparison Experiments}

We conduct comprehensive comparison experiments from three aspects to demonstrate the effectiveness of our proposed method.

\textbf{Effectiveness as a plug-and-play knowledge distillation strategy.} We first evaluate how our \proposedmethod enhances various knowledge distillation baselines on CUB-200, as shown in Table~\ref{tab:compare_cub200}. By integrating \proposedmethod into five representative KD methods (KD, RKD, SP, CATKD, TeKAP), we observe consistent performance improvements across all baselines with different teacher-student pairs. The gains range from 0.18\% to 4.49\%, demonstrating the effectiveness of our approach as a universal enhancement module.

\begin{table}[h]
    \centering
    \scriptsize
    \setlength{\tabcolsep}{3.5pt}
    \renewcommand{\arraystretch}{1}
    \begin{tabular}{ccccccc}
        \toprule
        & Teacher & VGG13 & ResNet50 & ResNet50 & ResNet50 & ResNet50 \\
        \midrule
        & Student & VGG8 & ResNet18 & ShuffleNetV2 & VGG8 & WRN-40-2 \\
        \midrule

        & KD & 39.50 & 65.20 & 64.44 & 41.49 & 53.08 \\
        \rowcolor{pink!20} & KD+ours & \textbf{40.73} & \textbf{67.05} & \textbf{65.81} & \textbf{42.68} & \textbf{54.45} \\
        & $\Delta$ & \textcolor{green!60!black}{1.23$\uparrow$} & \textcolor{green!60!black}{1.85$\uparrow$} & \textcolor{green!60!black}{1.37$\uparrow$} & \textcolor{green!60!black}{1.19$\uparrow$} & \textcolor{green!60!black}{1.37$\uparrow$} \\
        \midrule

        & RKD & 38.76 & 64.27 & 63.53 & 38.76 & 51.74 \\
        \rowcolor{pink!20} & RKD+ours & \textbf{39.33} & \textbf{65.21} & \textbf{64.92} & \textbf{41.16} & \textbf{52.30} \\
        & $\Delta$ & \textcolor{green!60!black}{0.57$\uparrow$} & \textcolor{green!60!black}{0.94$\uparrow$} & \textcolor{green!60!black}{1.39$\uparrow$} & \textcolor{green!60!black}{2.40$\uparrow$} & \textcolor{green!60!black}{0.56$\uparrow$} \\
        \midrule

        & SP & 38.81 & 63.29 & 64.00 & 38.95 & 52.67 \\
        \rowcolor{pink!20} & SP+ours & \textbf{39.66} & \textbf{65.36} & \textbf{64.96} & \textbf{41.61} & \textbf{53.21} \\
        & $\Delta$ & \textcolor{green!60!black}{0.85$\uparrow$} & \textcolor{green!60!black}{2.07$\uparrow$} & \textcolor{green!60!black}{0.96$\uparrow$} & \textcolor{green!60!black}{2.66$\uparrow$} & \textcolor{green!60!black}{0.54$\uparrow$} \\
        \midrule

        & CATKD & 38.79 & 63.32 & 63.88 & 39.85 & 52.69 \\
        \rowcolor{pink!20} & CATKD+ours & \textbf{39.61} & \textbf{67.19} & \textbf{64.98} & \textbf{44.34} & \textbf{53.21} \\
        & $\Delta$ & \textcolor{green!60!black}{0.82$\uparrow$} & \textcolor{green!60!black}{3.87$\uparrow$} & \textcolor{green!60!black}{1.10$\uparrow$} & \textcolor{green!60!black}{4.49$\uparrow$} & \textcolor{green!60!black}{0.52$\uparrow$} \\
        \midrule

        & TeKAP & 38.79 & 65.37 & 64.94 & 42.35 & 53.52 \\
        \rowcolor{pink!20} & TeKAP+ours & \textbf{39.63} & \textbf{65.88} & \textbf{65.12} & \textbf{43.41} & \textbf{54.11} \\
        & $\Delta$ & \textcolor{green!60!black}{0.84$\uparrow$} & \textcolor{green!60!black}{0.51$\uparrow$} & \textcolor{green!60!black}{0.18$\uparrow$} & \textcolor{green!60!black}{1.06$\uparrow$} & \textcolor{green!60!black}{0.59$\uparrow$} \\
        \bottomrule
    \end{tabular}
        \caption{Top-1 accuracy (\%) on CUB-200 with both homogeneous and heterogeneous teacher--student pairs. $\Delta$ indicates the accuracy gain when our method is applied.}
    \label{tab:compare_cub200}
\end{table}

\textbf{Effectiveness across diverse datasets.} We compare our approach (Ours+TeKAP) against five SOTA KD methods on four datasets, as presented in Table~\ref{tab:ours_vs_baselines}. Our method achieves the best performance on CIFAR-100 (74.50
\%), RAF-DB (87.45\%), DTD (68.29\%), and Stanford Dogs (66.85\%), outperforming all baseline methods. These consistent improvements across general object classification (CIFAR-100), facial expression recognition (RAF-DB), texture classification (DTD), and fine-grained dog breed recognition (Stanford Dogs) demonstrate the effectiveness and versatility of our multi-view fusion strategy in handling various visual recognition tasks.


\begin{table}[h]
\scriptsize
\setlength{\tabcolsep}{3pt}
\renewcommand{\arraystretch}{1.15}
\centering
\resizebox{\columnwidth}{!}{%
\begin{tabular}{l l l c c c c c c}
\toprule
Dataset & Teacher & Student & KD & RKD & SP & CATKD & TeKAP & \cellcolor{pink!20}Ours \\
\midrule
CIFAR100 & 75.74 & 71.99 & 74.28 & 73.36 & 72.08 & 72.11 & \underline{74.39} & \cellcolor{pink!20}\textbf{74.50} \\
RAF-DB   & 86.96 & 81.88 & 85.49 & \underline{86.96} & 84.64 & 84.68 & 86.31 & \cellcolor{pink!20}\textbf{87.45} \\
DTD      & 68.83 & 66.44 & 66.86 & \underline{68.08} & 66.75 & 62.40 & 66.75 & \cellcolor{pink!20}\textbf{68.29} \\
SFDOGS   & 68.44 & 61.88 & 66.30 & 63.79 & 62.37 & 62.40 & \underline{66.62} & \cellcolor{pink!20}\textbf{66.85} \\
\bottomrule
\end{tabular}}
\caption{Comparison with state-of-the-art KD baselines. Best results are in \textbf{bold}, and second-best are \underline{underlined}.}
\label{tab:ours_vs_baselines}
\end{table}

\textbf{Effectiveness across different architectures.} We further examine our framework's performance with various student architectures, as shown in Table~\ref{tab:placeholder}. When integrated into CATKD and TeKAP, our method consistently improves all evaluation metrics for both ResNet18 and VGG8 students. For ResNet18, we achieve gains of 2.07\%/0.50\% in Top-1 accuracy when enhancing CATKD/TeKAP respectively. More notably, the lightweight VGG8 architecture shows even larger improvements of 4.49\%/1.06\% in Top-1 accuracy, along with substantial gains in Top-5 accuracy and recall. These results demonstrate that our multi-view fusion strategy is particularly effective for smaller student models with limited capacity, making it especially valuable for resource-constrained deployment scenarios.

\begin{table}[h]
\scriptsize
\renewcommand{\arraystretch}{1.2}  
\setlength{\tabcolsep}{3pt}        
\centering
\begin{tabularx}{\columnwidth}{c *{3}{>{\centering\arraybackslash}X}|*{3}{>{\centering\arraybackslash}X}}
\hline
Teacher & \multicolumn{6}{c}{ResNet50} \\
\hline
Student & \multicolumn{3}{c}{ResNet18} & \multicolumn{3}{c}{VGG8} \\
\hline
Metric & Top-1 & Top-5 & Recall & Top-1 & Top-5 & Recall \\
\hline
CATKD & 63.29 & 85.00 & 63.37 & 39.85 & 64.20 & 39.05 \\
\rowcolor{pink!20} \textbf{CATKD+ours} & \textbf{65.36} & \textbf{86.88} & \textbf{65.79} & \textbf{44.34} & \textbf{67.71} & \textbf{43.23} \\
TeKAP & 65.38 & 86.90 & 65.46 & 42.35 & 68.80 & 42.62 \\
\rowcolor{pink!20} \textbf{TeKAP+ours} & \textbf{65.88} & \textbf{87.23} & \textbf{66.00} & \textbf{43.41} & \textbf{70.33} & \textbf{43.69} \\
\hline
\end{tabularx}
\caption{Evaluation of the generalization ability of our method across different teacher–student pairs and baselines. Best results are in \textbf{bold}. }
\label{tab:placeholder}
\end{table}

\subsection{Ablation Study}
We conduct ablation experiments on CUB-200 to evaluate different view combinations and fusion strategies. As shown in Table~\ref{tab:model_ablation}, adding edge or high-frequency views to RGB individually brings +1.59\% improvement. However, simply averaging all three views yields only +0.40\% gain, while our adaptive fusion achieves +2.82\% improvement. This demonstrates that both multi-view inputs and proper fusion mechanisms are essential for effective knowledge distillation.

\newcolumntype{C}{>{\centering\arraybackslash}X}  

\begin{table}[t]
    \scriptsize
    \centering
    \renewcommand{\arraystretch}{1.2}
    \begin{tabularx}{\columnwidth}{lCC}  
        \toprule
        \textbf{Modality Configuration} & \textbf{Acc (\%)} & $\boldsymbol{\Delta}$ \\
        \midrule
        RGB & 41.52 & -- \\
        RGB + Edge & 43.11 & +1.59 \\
        RGB + High-frequency & 43.11 & +1.59 \\
        Edge + High-frequency & 41.92 & +0.40 \\
        \rowcolor{pink!20} RGB + Edge + High-frequency & \textbf{44.34} & +2.82 \\
        \bottomrule
    \end{tabularx}
    \caption{Ablation study of multiple image views. Best results are in \textbf{bold}.}
    \label{tab:model_ablation}
\end{table}

Table~\ref{tab:loss_ablation} presents the ablation of different distillation losses. While using only logits loss achieves 65.20\%, adding feature-level or CRD loss individually brings marginal improvements (+0.05\% and +0.17\%). Combining all three losses yields the best accuracy of 65.40\%, confirming that each loss provides complementary supervision for effective knowledge transfer.

\newcolumntype{C}{>{\centering\arraybackslash}X}  

\begin{table}[h]
    \centering
    \scriptsize
    \renewcommand{\arraystretch}{1.2}
    \begin{tabularx}{\columnwidth}{CCC C}  
    \toprule
    $\bm{\mathcal{L}}_{\mathrm{logits}}$ &
    $\bm{\mathcal{L}}_{\mathrm{feat}}$ &
    $\bm{\mathcal{L}}_{\mathrm{CRD}}$ &
    \textbf{Acc (\%)} \\
    \midrule

        \checkmark &  &  & 65.20 \\
        \checkmark & \checkmark &  & 65.25 \\
        \checkmark &  & \checkmark & 65.37 \\
        \rowcolor{pink!20} \checkmark & \checkmark & \checkmark & \textbf{65.40} \\
        \bottomrule
    \end{tabularx}
    \caption{Ablation study of different distillation loss terms. Best results are in \textbf{bold}.}
    \label{tab:loss_ablation}
\end{table}

\subsection{Visualization Analysis}
\begin{figure}[h]
    \centering
    \includegraphics[width=1\linewidth]{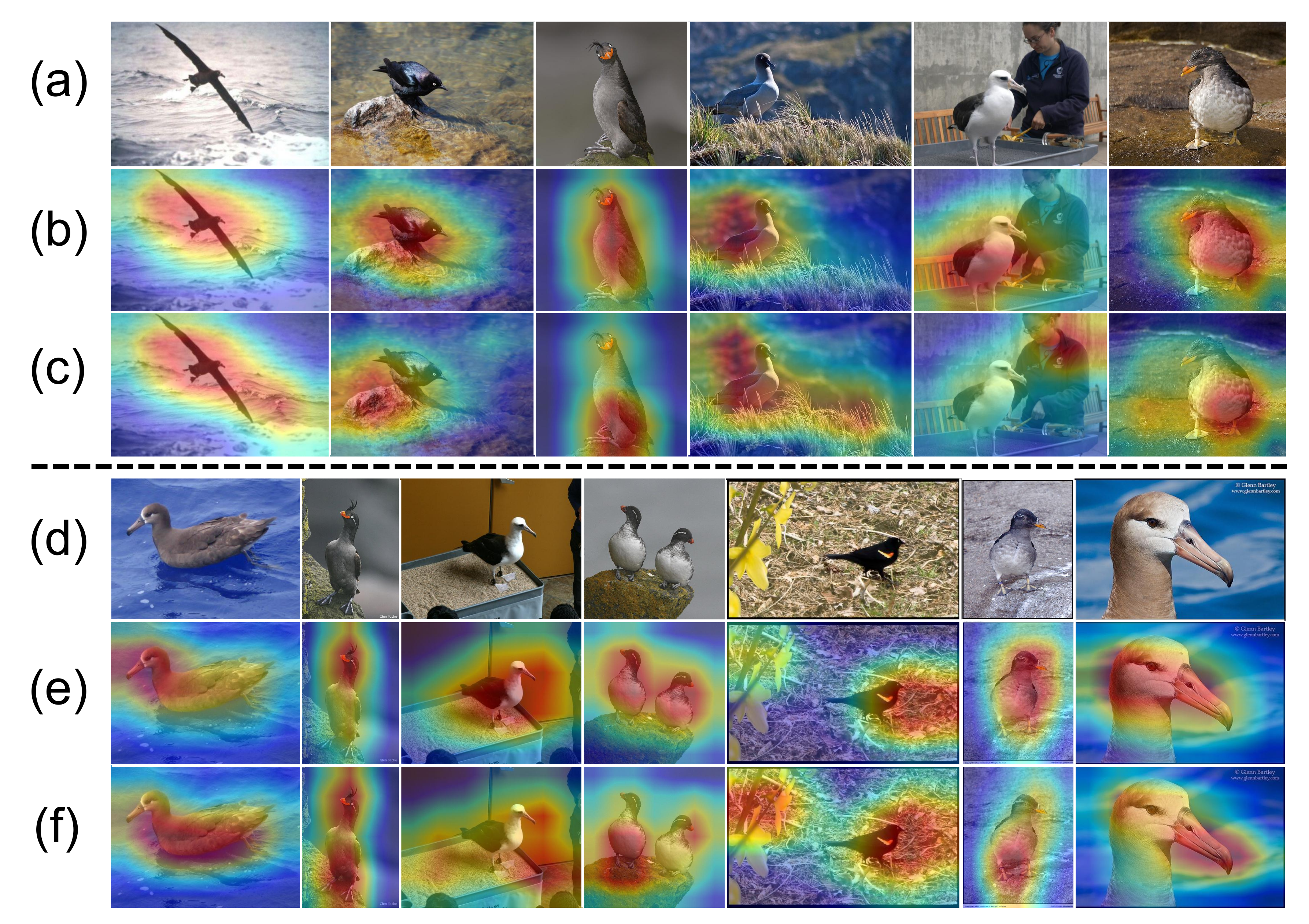}
    \caption{Attention heatmap visualization.
(a)(d) Input image.
(b)(e) Attention map of our proposed method.
(c)(f) Attention map of the baseline method CAT-KD.
}
    \label{fig:heatmap}
\end{figure}

We visualize attention maps  in Fig.~\ref{fig:heatmap}. Our method (row b, e) produces more focused and semantically coherent attention regions compared to baseline CAT-KD (row c, f). Specifically, our method precisely localizes discriminative object parts (e.g., bird heads and distinctive body features) while effectively suppressing background noise.
In contrast, CAT-KD generates scattered and diffused attention patterns, activating irrelevant background regions. These visualization results confirm that our multi-view distillation strategy enhances the student's ability to localize semantically relevant features.

\begin{figure*}[h]  
    \centering
    \includegraphics[width=\textwidth]{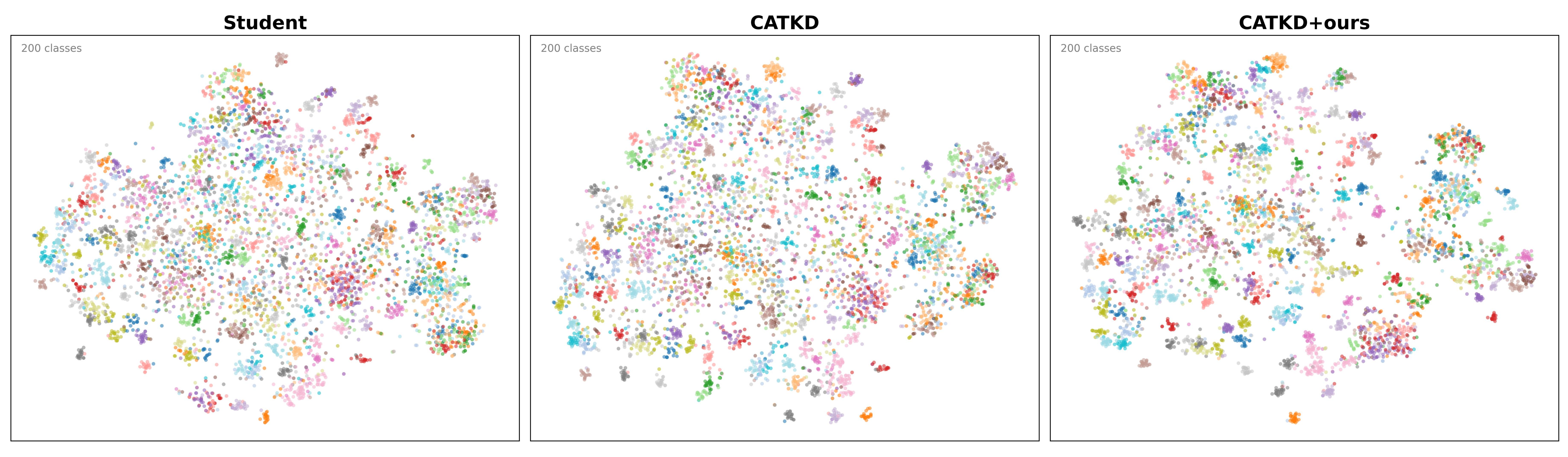}  
    \caption{Scatter plots of Student, CAT-KD, and CAT-KD+ours on CUB-200. 
    Our method combined with CAT-KD generates tighter and more separated clusters, indicating improved class-wise consistency and feature discriminability.}
    \label{fig:tsne}
\end{figure*}

We visualize feature distributions using t-SNE on CUB-200 datase  in Fig.~\ref{fig:tsne}. Our method (CAT-KD+ours) produces tighter, more separated clusters compared to baseline CAT-KD, demonstrating improved intra-class cohesion and inter-class discrimination. This enhanced feature organization validates that our multi-view enhancement effectively improves knowledge transfer quality.

\begin{figure}[h]
    \centering
    \includegraphics[width=1\linewidth]{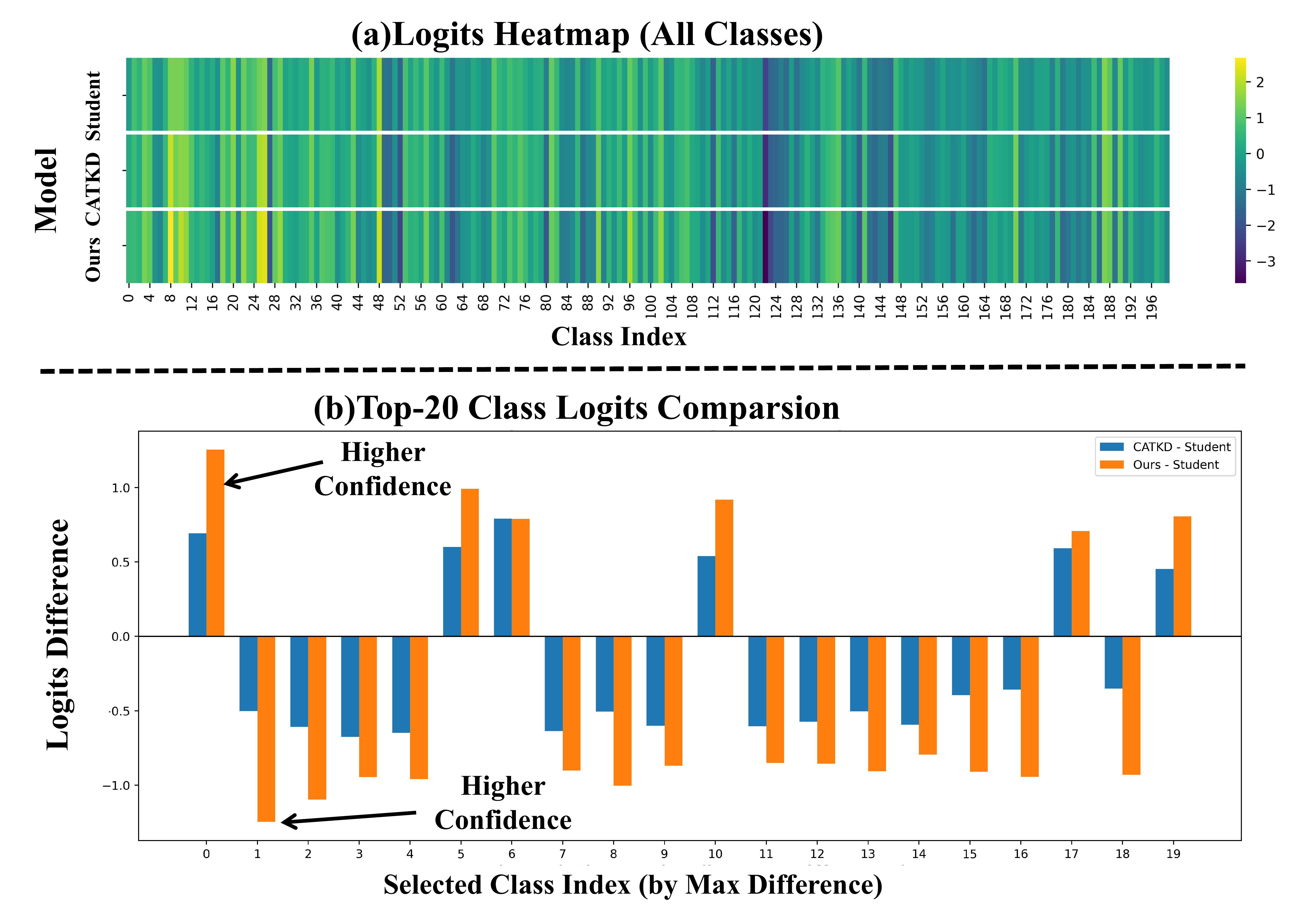}
    \caption{Class-wise logit heatmaps and distribution.}
    \label{fig:logits_pixl}
\end{figure}

We analyze class-wise logit distributions to evaluate prediction quality in Fig.~\ref{fig:logits_pixl}. The heatmap (a) shows logit values across all classes, while (b) compares top-20 class predictions between CAT-KD baseline and our method. Our method produces sharper logit peaks with higher confidence for correct classes and better suppression of incorrect ones. This enhanced class separation demonstrates that our multi-view framework effectively transfers both class-specific and inter-class relational knowledge, yielding more discriminative predictions.

\begin{figure}[h]
    \centering
    \includegraphics[width=1\linewidth]{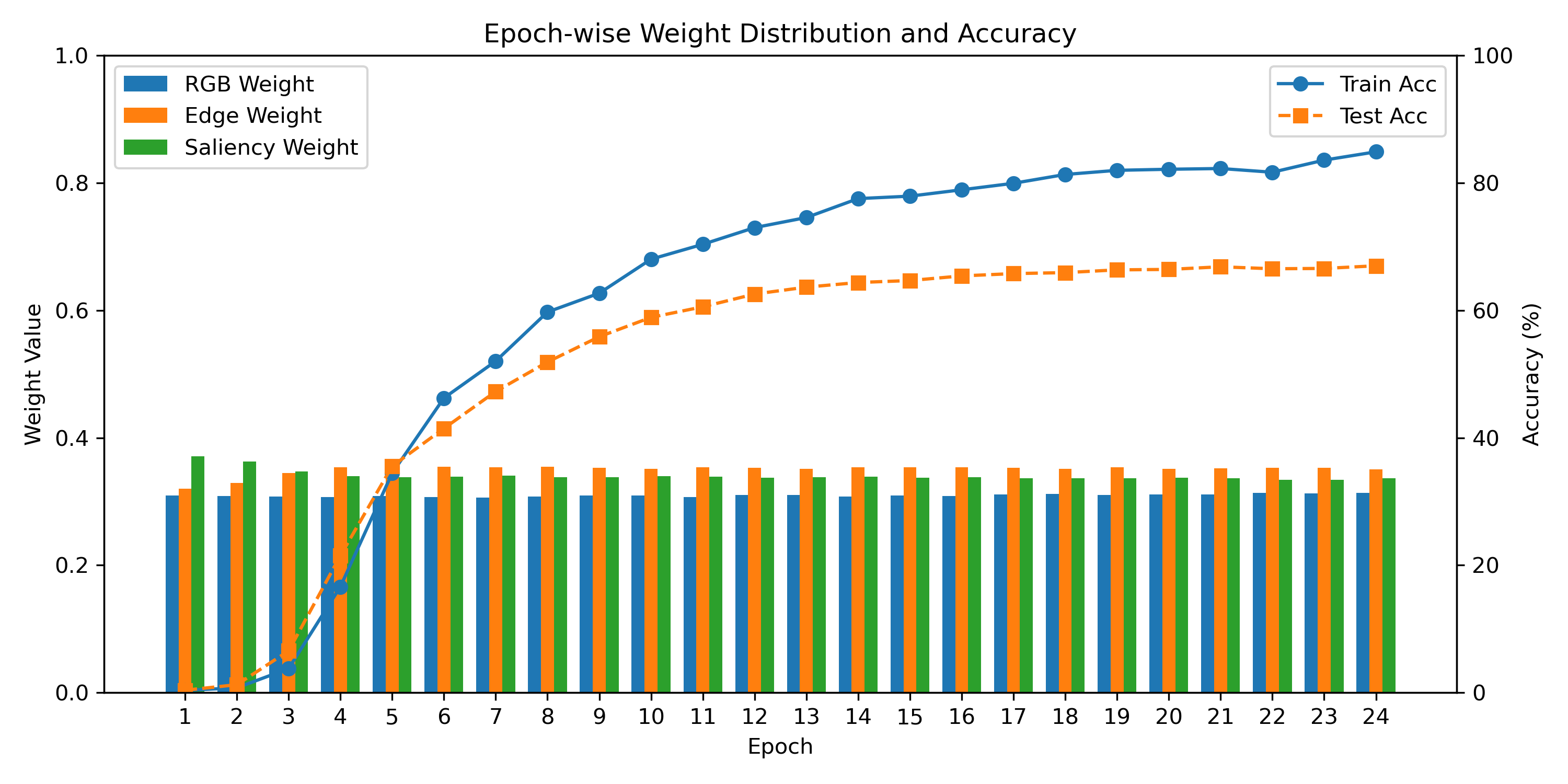}
    \caption{Epoch-wise feature fusion weights (RGB, edge, high-frequency) and corresponding accuracy. 
Our adaptive weighting mechanism dynamically adjusts modality importance, leading to improved training and test performance.}
    \label{fig:line_bar_chart}
\end{figure}

To quantitatively analyze the importance of each visual view during feature fusion, we track the evolution of adaptive fusion weights and accuracy throughout training. The visualization results are shown in Fig.~\ref{fig:line_bar_chart}. Initially, high-frequency features receive higher weights (~0.37) compared to RGB (0.31) and edge (0.32), indicating their dominant role in early learning. As training progresses, the model adaptively rebalances modality contributions, with high-frequency weights gradually decreasing while RGB and edge weights stabilize. This dynamic adjustment correlates strongly with accuracy improvement, as test accuracy rises from 0.38\% to 66.98\% by epoch 24. The synchronized convergence of feature weights and accuracy gains confirms that our adaptive weighting mechanism effectively integrates complementary modalities, enabling more robust and discriminative representation learning.

\section{Conclusion}

We propose a text-guided multi-view knowledge distillation framework \proposedmethod that leverages dual-modality teachers—a visual teacher and a text teacher (CLIP)—to provide enriched supervisory signals. We enhance the visual teacher with multi-view inputs incorporating edge and high-frequency priors, which are adaptively fused using CLIP-generated semantic weights. Through feature-level and logits-level distillation from the visual teacher, combined with vision-language contrastive regularization from the text teacher, our method strengthens semantic knowledge in the student model. Extensive experiments on five benchmarks demonstrate that \proposedmethod consistently improves existing KD methods by up to 4.49\%, confirming that integrating complementary visual priors with semantic guidance significantly enhances knowledge distillation. Future work will explore extending this framework to other visual tasks and investigating adaptive view generation strategies.
\appendix

\bibliographystyle{named}
\bibliography{ijcai26}

@article{hinton2015distilling,
  author       = {Geoffrey E. Hinton and
                  Oriol Vinyals and
                  Jeffrey Dean},
  title        = {Distilling the Knowledge in a Neural Network},
  journal      = {CoRR},
  volume       = {abs/1503.02531},
  year         = {2015},
  url          = {http://arxiv.org/abs/1503.02531},
  eprinttype    = {arXiv},
  eprint       = {1503.02531},

}

@article{gou2021knowledge,
  author       = {Jianping Gou and
                  Baosheng Yu and
                  Stephen J. Maybank and
                  Dacheng Tao},
  title        = {Knowledge Distillation: A Survey},
  journal      = {International Journal of Computer Vision},
  volume       = {129},
  number       = {6},
  pages        = {1789--1819},
  year         = {2021},
  url          = {https://doi.org/10.1007/s11263-021-01453-z},

}

@article{SunWHC2025,
  author       = {Yafeng Sun and
                  Xingwang Wang and
                  Junhong Huang and
                  Shilin Chen},
  title        = {Balanced sample repository for knowledge distillation in data-free
                  image classification scenario},
  journal      = {Eng. Appl. Artif. Intell.},
  volume       = {161},
  pages        = {112160},
  year         = {2025},
  url          = {https://doi.org/10.1016/j.engappai.2025.112160},
  doi          = {10.1016/J.ENGAPPAI.2025.112160},

}

@article{BaiLKYY2026,
  author       = {Yaping Bai and
                  Jinghua Li and
                  Dehui Kong and
                  Suqiao Yang and
                  Baocai Yin},
  title        = {{EKDSC:} Long-tailed recognition based on expert knowledge distillation
                  for specific categories},
  journal      = {Neural Networks},
  volume       = {194},
  pages        = {108099},
  year         = {2026},
  url          = {https://doi.org/10.1016/j.neunet.2025.108099},
  doi          = {10.1016/J.NEUNET.2025.108099},

}

@article{Yang2025edge,
  author       = {Jinxin Yang and
                  Wujie Zhou},
  title        = {Cross-attention fusion and edge-guided fully supervised contrastive
                  learning network for rail surface defect detection},
  journal      = {Appl. Intell.},
  volume       = {55},
  number       = {6},
  pages        = {421},
  year         = {2025},
  url          = {https://doi.org/10.1007/s10489-025-06314-7},
  doi          = {10.1007/S10489-025-06314-7},

}

@article{Cantone2026sal_guide,
  author       = {Marco Cantone and
                  Ciro Russo and
                  Federico V. L. Dell'Ascenza and
                  Claudio Marrocco and
                  Alessandro Bria},
  title        = {Deep learning for {DBT} classification with saliency-guided 2D synthesis},
  journal      = {Pattern Recognit.},
  volume       = {172},
  pages        = {112316},
  year         = {2026},
  url          = {https://doi.org/10.1016/j.patcog.2025.112316},
  doi          = {10.1016/J.PATCOG.2025.112316},

}

@inproceedings{park2019relational,
  author       = {Wonpyo Park and
                  Dongju Kim and
                  Yoonho Lu and
                  Minsu Cho},
  title        = {Relational Knowledge Distillation},
  booktitle    = {Proceedings of the IEEE/CVF Conference on Computer Vision and Pattern Recognition (CVPR)},
  pages        = {3967--3976},
  year         = {2019},
  publisher    = {IEEE},

}

@inproceedings{CLIP,
  author       = {Alec Radford and
                  Jong Wook Kim and
                  Chris Hallacy and
                  Aditya Ramesh and
                  Gabriel Goh and
                  Sandhini Agarwal and
                  Girish Sastry and
                  Amanda Askell and
                  Pamela Mishkin and
                  Jack Clark and
                  Gretchen Krueger and
                  Ilya Sutskever},
  editor       = {Marina Meila and
                  Tong Zhang},
  title        = {Learning Transferable Visual Models From Natural Language Supervision},
  booktitle    = {Proceedings of the 38th International Conference on Machine Learning},
  series       = {Proceedings of Machine Learning Research},
  volume       = {139},
  pages        = {8748--8763},
  publisher    = {{PMLR}},
  year         = {2021},
  url          = {http://proceedings.mlr.press/v139/radford21a.html},

}

@article{Zhao2022DKD,
  author       = {Borui Zhao and
                  Quan Cui and
                  Renjie Song and
                  Yiyu Qiu and
                  Jiajun Liang},
  title        = {Decoupled Knowledge Distillation},
  journal      = {CoRR},
  volume       = {abs/2203.08679},
  year         = {2022},
  url          = {https://doi.org/10.48550/arXiv.2203.08679},
  doi          = {10.48550/ARXIV.2203.08679},
  eprinttype    = {arXiv},
  eprint       = {2203.08679},

}

@inproceedings{Zi2023catkd,
  author       = {Ziyao Guo and
                  Haonan Yan and
                  Hui Li and
                  Xiaodong Lin},
  title        = {Class Attention Transfer Based Knowledge Distillation},
  booktitle    = {{IEEE/CVF} Conference on Computer Vision and Pattern Recognition},
  pages        = {11868--11877},
  publisher    = {{IEEE}},
  year         = {2023},
  url          = {https://doi.org/10.1109/CVPR52729.2023.01142},
  doi          = {10.1109/CVPR52729.2023.01142},

}

@inproceedings{Tung2019SPKD,
  author       = {Frederick Tung and
                  Greg Mori},
  title        = {Similarity-Preserving Knowledge Distillation},
  booktitle    = {2019 {IEEE/CVF} International Conference on Computer Vision},
  pages        = {1365--1374},
  publisher    = {{IEEE}},
  year         = {2019},
  url          = {https://doi.org/10.1109/ICCV.2019.00145},
  doi          = {10.1109/ICCV.2019.00145},

}

@inproceedings{Jang2025VL2Lite,
  author       = {Jinseong Jang and
                  Chunfei Ma and
                  Byeongwon Lee},
  title        = {VL2Lite: Task-Specific Knowledge Distillation from Large Vision-Language
                  Models to Lightweight Networks},
  booktitle    = {{IEEE/CVF} Conference on Computer Vision and Pattern Recognition},
  pages        = {30073--30083},
  publisher    = {Computer Vision Foundation / {IEEE}},
  year         = {2025},
  url          = {https://openaccess.thecvf.com/content/CVPR2025/html/Jang\_VL2Lite\_Task-Specific\_Knowledge\_Distillation\_from\_Large\_Vision-Language\_Models\_to\_Lightweight\_CVPR\_2025\_paper.html},
  doi          = {10.1109/CVPR52734.2025.02799},

}

@inproceedings{Ta2025MSCLIPKD,
  author       = {Taegyeong Lee and
                  Jinsik Bang and
                  Soyeong Kwon and
                  Taehwan Kim},
  title        = {Multi-aspect Knowledge Distillation with Large Language Model},
  booktitle    = {{IEEE/CVF} Conference on Computer Vision and Pattern Recognition Workshops},
  pages        = {2121--2130},
  publisher    = {Computer Vision Foundation / {IEEE}},
  year         = {2025},
  url          = {https://openaccess.thecvf.com/content/CVPR2025W/FGVC/html/Lee\_Multi-aspect\_Knowledge\_Distillation\_with\_Large\_Language\_Model\_CVPRW\_2025\_paper.html},

}

@inproceedings{Zhou2025CLIPIKD,
  author       = {Jingtao Zhou and
                  Hao Zheng and
                  Wenkai Zhong and
                  Zhiqiang Bao},
  title        = {Improving Knowledge Distillation via Cross-Modal Insights from {CLIP}},
  booktitle    = {2025 {IEEE} International Conference on Acoustics, Speech and Signal
                  Processing},
  pages        = {1--5},
  publisher    = {{IEEE}},
  year         = {2025},
  url          = {https://doi.org/10.1109/ICASSP49660.2025.10889763},
  doi          = {10.1109/ICASSP49660.2025.10889763},

}

@inproceedings{Md2025TeKAP,
  author       = {Md. Imtiaz Hossain and
                  Sharmen Akhter and
                  Choong Seon Hong and
                  Eui{-}Nam Huh},
  title        = {Single Teacher, Multiple Perspectives: Teacher Knowledge Augmentation
                  for Enhanced Knowledge Distillation},
  booktitle    = {The Thirteenth International Conference on Learning Representations},
  publisher    = {OpenReview.net},
  year         = {2025},
  url          = {https://openreview.net/forum?id=DmEHmZ89iB},

}

@inproceedings{Yang2025MTKD,
  author       = {Chuanguang Yang and
                  Xinqiang Yu and
                  Han Yang and
                  Zhulin An and
                  Chengqing Yu and
                  Libo Huang and
                  Yongjun Xu},
  editor       = {Toby Walsh and
                  Julie Shah and
                  Zico Kolter},
  title        = {Multi-Teacher Knowledge Distillation with Reinforcement Learning for
                  Visual Recognition},
  booktitle    = {AAAI-25, Sponsored by the Association for the Advancement of Artificial
                  Intelligence},
  pages        = {9148--9156},
  publisher    = {{AAAI} Press},
  year         = {2025},
  url          = {https://doi.org/10.1609/aaai.v39i9.32990},
  doi          = {10.1609/AAAI.V39I9.32990},

}

@inproceedings{Zhang2023MMKD,
  author       = {Hailin Zhang and
                  Defang Chen and
                  Can Wang},
  title        = {Adaptive Multi-Teacher Knowledge Distillation with Meta-Learning},
  booktitle    = {{IEEE} International Conference on Multimedia and Expo},
  pages        = {1943--1948},
  publisher    = {{IEEE}},
  year         = {2023},
  url          = {https://doi.org/10.1109/ICME55011.2023.00333},
  doi          = {10.1109/ICME55011.2023.00333},

}

@inproceedings{Jiang2024MTKD,
  author       = {Yuxuan Jiang and
                  Chen Feng and
                  Fan Zhang and
                  David Bull},
  editor       = {Ales Leonardis and
                  Elisa Ricci and
                  Stefan Roth and
                  Olga Russakovsky and
                  Torsten Sattler and
                  G{\"{u}}l Varol},
  title        = {{MTKD:} Multi-Teacher Knowledge Distillation for Image Super-Resolution},
  booktitle    = {Computer Vision - {ECCV} 2024},
  series       = {Lecture Notes in Computer Science},
  volume       = {15097},
  pages        = {364--382},
  publisher    = {Springer},
  year         = {2024},
  url          = {https://doi.org/10.1007/978-3-031-72933-1\_21},
  doi          = {10.1007/978-3-031-72933-1\_21},

}

@article{Ma2024CAGMKD,
  author       = {Dongtong Ma and
                  Kaibing Zhang and
                  Qizhi Cao and
                  Jie Li and
                  Xinbo Gao},
  title        = {Coordinate Attention Guided Dual-Teacher Adaptive Knowledge Distillation
                  for image classification},
  journal      = {Expert Syst. Appl.},
  volume       = {250},
  pages        = {123892},
  year         = {2024},
  url          = {https://doi.org/10.1016/j.eswa.2024.123892},
  doi          = {10.1016/J.ESWA.2024.123892},

}

@inproceedings{Liu2025NNMTKD,
  author       = {Yujing Liu and
                  Zongqian Wu and
                  Zhengyu Lu and
                  Ci Nie and
                  Guoqiu Wen and
                  Yonghua Zhu and
                  Xiaofeng Zhu},
  editor       = {Toby Walsh and
                  Julie Shah and
                  Zico Kolter},
  title        = {Noisy Node Classification by Bi-level Optimization Based Multi-Teacher
                  Distillation},
  booktitle    = {AAAI-25, Sponsored by the Association for the Advancement of Artificial
                  Intelligence},
  pages        = {19033--19040},
  publisher    = {{AAAI} Press},
  year         = {2025},
  url          = {https://doi.org/10.1609/aaai.v39i18.34095},
  doi          = {10.1609/AAAI.V39I18.34095},

}

@inproceedings{Wu2025MRSKD,
  author       = {Shuang Wu and
                  Heng Liang and
                  Yong Zhang and
                  Yanlin Chen and
                  Ziyu Jia},
  title        = {A Cross-Modal Densely Guided Knowledge Distillation Based on Modality
                  Rebalancing Strategy for Enhanced Unimodal Emotion Recognition},
  booktitle    = {Proceedings of the Thirty-Fourth International Joint Conference on
                  Artificial Intelligence},
  pages        = {4236--4244},
  publisher    = {ijcai.org},
  year         = {2025},
  url          = {https://doi.org/10.24963/ijcai.2025/472},
  doi          = {10.24963/IJCAI.2025/472},

}

@inproceedings{Liu2024LLMKD,
  author       = {Chen Liu and
                  Shibo He and
                  Qihang Zhou and
                  Shizhong Li and
                  Wenchao Meng},
  title        = {Large Language Model Guided Knowledge Distillation for Time Series
                  Anomaly Detection},
  booktitle    = {Proceedings of the Thirty-Third International Joint Conference on
                  Artificial Intelligence},
  pages        = {2162--2170},
  publisher    = {ijcai.org},
  year         = {2024},
  url          = {https://www.ijcai.org/proceedings/2024/239},

}

@inproceedings{Zhao2025BRSKD,
  author       = {Di Zhao and
                  Jingfeng Zhang and
                  Hongsheng Hu and
                  Philippe Fournier{-}Viger and
                  Gillian Dobbie and
                  Yun Sing Koh},
  title        = {Balancing Invariant and Specific Knowledge for Domain Generalization
                  with Online Knowledge Distillation},
  booktitle    = {Proceedings of the Thirty-Fourth International Joint Conference on
                  Artificial Intelligence},
  pages        = {2440--2448},
  publisher    = {ijcai.org},
  year         = {2025},
  url          = {https://doi.org/10.24963/ijcai.2025/272},
  doi          = {10.24963/IJCAI.2025/272},

}

@inproceedings{Mora2024FLKD,
  author       = {Alessio Mora and
                  Irene Tenison and
                  Paolo Bellavista and
                  Irina Rish},
  title        = {Knowledge Distillation in Federated Learning: {A} Practical Guide},
  booktitle    = {Proceedings of the Thirty-Third International Joint Conference on
                  Artificial Intelligence},
  pages        = {8188--8196},
  publisher    = {ijcai.org},
  year         = {2024},
  url          = {https://www.ijcai.org/proceedings/2024/905},

}

@article{Cai2026ODKD,
  author       = {Bo Cai and
                  Houjie Li and
                  Yanping Yang and
                  Jin Yan},
  title        = {CFF-KDNet: Cross-scale feature fusion network with knowledge distillation
                  for camouflaged object detection},
  journal      = {Expert Syst. Appl.},
  volume       = {299},
  pages        = {130209},
  year         = {2026},
  url          = {https://doi.org/10.1016/j.eswa.2025.130209},
  doi          = {10.1016/J.ESWA.2025.130209},

}

@article{Zhang2024selfkd,
  author       = {Xin Zhang and
                  Jinlin Zhu and
                  Dongjing Wang and
                  Yueyun Wang and
                  Tingting Liang and
                  Hongbo Wang and
                  Yuyu Yin},
  title        = {A gradual self distillation network with adaptive channel attention
                  for facial expression recognition},
  journal      = {Appl. Soft Comput.},
  volume       = {161},
  pages        = {111762},
  year         = {2024},
  url          = {https://doi.org/10.1016/j.asoc.2024.111762},
  doi          = {10.1016/J.ASOC.2024.111762},
  bibsource    = {dblp computer science bibliography, https://dblp.org}
}

@article{Yong2019CRD,
  author       = {Yonglong Tian and
                  Dilip Krishnan and
                  Phillip Isola},
  title        = {Contrastive Representation Distillation},
  journal      = {CoRR},
  volume       = {abs/1910.10699},
  year         = {2019},
  url          = {http://arxiv.org/abs/1910.10699},
  eprinttype    = {arXiv},
  eprint       = {1910.10699},
  bibsource    = {dblp computer science bibliography, https://dblp.org}
}

@inproceedings{Wang2025SeqMvRL,
  author       = {Ren Wang and
                  Haoliang Sun and
                  Yuxiu Lin and
                  Chuanhui Zuo and
                  Yongshun Gong and
                  Yilong Yin and
                  Wenjia Meng},
  title        = {SeqMvRL: {A} Sequential Fusion Framework for Multi-view Representation
                  Learning},
  booktitle    = {{IEEE/CVF} Conference on Computer Vision and Pattern Recognition},
  pages        = {25822--25831},
  publisher    = {Computer Vision Foundation / {IEEE}},
  year         = {2025},
  url          = {https://openaccess.thecvf.com/content/CVPR2025/html/Wang\_SeqMvRL\_A\_Sequential\_Fusion\_Framework\_for\_Multi-view\_Representation\_Learning\_CVPR\_2025\_paper.html},
  doi          = {10.1109/CVPR52734.2025.02405},
  bibsource    = {dblp computer science bibliography, https://dblp.org}
}

@inproceedings{Bao2024MVF,
  author       = {Yiwei Bao and
                  Feng Lu},
  title        = {Unsupervised Gaze Representation Learning from Multi-view Face Images},
  booktitle    = {{IEEE/CVF} Conference on Computer Vision and Pattern Recognition},
  pages        = {1419--1428},
  publisher    = {{IEEE}},
  year         = {2024},
  url          = {https://doi.org/10.1109/CVPR52733.2024.00141},
  doi          = {10.1109/CVPR52733.2024.00141},
  bibsource    = {dblp computer science bibliography, https://dblp.org}
}

@inproceedings{NiZ23IJCAIKD,
  author       = {Zhenliang Ni and
                  Fukui Yang and
                  Shengzhao Wen and
                  Gang Zhang},
  title        = {Dual Relation Knowledge Distillation for Object Detection},
  booktitle    = {Proceedings of the Thirty-Second International Joint Conference on
                  Artificial Intelligence},
  pages        = {1276--1284},
  publisher    = {ijcai.org},
  year         = {2023},
  url          = {https://doi.org/10.24963/ijcai.2023/142},
  doi          = {10.24963/IJCAI.2023/142},
  bibsource    = {dblp computer science bibliography, https://dblp.org}
}

@inproceedings{Lee2025cvprkd,
  author       = {Taegyeong Lee and
                  Jinsik Bang and
                  Soyeong Kwon and
                  Taehwan Kim},
  title        = {Multi-aspect Knowledge Distillation with Large Language Model},
  booktitle    = {{IEEE/CVF} Conference on Computer Vision and Pattern Recognition Workshops},
  pages        = {2121--2130},
  publisher    = {Computer Vision Foundation / {IEEE}},
  year         = {2025},
  url          = {https://openaccess.thecvf.com/content/CVPR2025W/FGVC/html/Lee\_Multi-aspect\_Knowledge\_Distillation\_with\_Large\_Language\_Model\_CVPRW\_2025\_paper.html},
  bibsource    = {dblp computer science bibliography, https://dblp.org}
}

@inproceedings{Chen2025VLM,
  author       = {Ziliang Chen and
                  Xin Huang and
                  Xiaoxuan Fan and
                  Keze Wang and
                  Yuyu Zhou and
                  Quanlong Guan and
                  Liang Lin},
  title        = {Reproducible Vision-Language Models Meet Concepts Out of Pre-Training},
  booktitle    = {{IEEE/CVF} Conference on Computer Vision and Pattern Recognition},
  pages        = {14701--14711},
  publisher    = {Computer Vision Foundation / {IEEE}},
  year         = {2025},
  url          = {https://openaccess.thecvf.com/content/CVPR2025/html/Chen\_Reproducible\_Vision-Language\_Models\_Meet\_Concepts\_Out\_of\_Pre-Training\_CVPR\_2025\_paper.html},
  doi          = {10.1109/CVPR52734.2025.01370},
  bibsource    = {dblp computer science bibliography, https://dblp.org}
}

@inproceedings{Chen2025LRME,
  author       = {Qizhou Chen and
                  Chengyu Wang and
                  Dakan Wang and
                  Taolin Zhang and
                  Wangyue Li and
                  Xiaofeng He},
  title        = {Lifelong Knowledge Editing for Vision Language Models with Low-Rank
                  Mixture-of-Experts},
  booktitle    = {{IEEE/CVF} Conference on Computer Vision and Pattern Recognition,
                  {CVPR} 2025, Nashville, TN, USA, June 11-15, 2025},
  pages        = {9455--9466},
  publisher    = {Computer Vision Foundation / {IEEE}},
  year         = {2025},
  url          = {https://openaccess.thecvf.com/content/CVPR2025/html/Chen\_Lifelong\_Knowledge\_Editing\_for\_Vision\_Language\_Models\_with\_Low-Rank\_Mixture-of-Experts\_CVPR\_2025\_paper.html},
  doi          = {10.1109/CVPR52734.2025.00883},
  bibsource    = {dblp computer science bibliography, https://dblp.org}
}

@inproceedings{Eutamene2024SIDVLM,
  title={Synthetic Image Detection Using Mixture of Knowledge Distillation from Vision-Language Models},
  author={Bougueffa Eutamene, Hessen and Hamidouche, Wassim and Keita, Mamadou and Taleb-Ahmed, Abdelmalik and Hadid, Abdenour},
  booktitle={International Conference on Pattern Recognition},
  pages={226--243},
  year={2024},
  organization={Springer}
}

@inproceedings{Li2022ISSKD,
  author       = {Maohui Li and
                  Michael Halstead and
                  Chris McCool},
  title        = {Knowledge Distillation for Efficient Instance Semantic Segmentation
                  with Transformers},
  booktitle    = {{IEEE/CVF} Conference on Computer Vision and Pattern Recognition},
  pages        = {5432--5439},
  publisher    = {{IEEE}},
  year         = {2024},
  url          = {https://doi.org/10.1109/CVPRW63382.2024.00552},
  doi          = {10.1109/CVPRW63382.2024.00552},
  bibsource    = {dblp computer science bibliography, https://dblp.org}
}

@inproceedings{Kong2025CLIPExtract,
  author       = {Fanxuan Kong and
                  Jun Lu},
  title        = {A Weakly Supervised Semantic Segmentation Model with Enhanced {CLIP}
                  Feature Extraction},
  booktitle    = {2025 {IEEE} International Conference on Acoustics},
  pages        = {1--5},
  publisher    = {{IEEE}},
  year         = {2025},
  url          = {https://doi.org/10.1109/ICASSP49660.2025.10888373},
  doi          = {10.1109/ICASSP49660.2025.10888373},
  bibsource    = {dblp computer science bibliography, https://dblp.org}
}

@inproceedings{Guo2023Speakerextract,
  author       = {Zihao Guo and
                  Shilin Wang},
  title        = {Content-Insensitive Dynamic Lip Feature Extraction for Visual Speaker
                  Authentication Against Deepfake Attacks},
  booktitle    = {{IEEE} International Conference on Acoustics},
  pages        = {1--5},
  publisher    = {{IEEE}},
  year         = {2023},
  url          = {https://doi.org/10.1109/ICASSP49357.2023.10096249},
  doi          = {10.1109/ICASSP49357.2023.10096249},
  bibsource    = {dblp computer science bibliography, https://dblp.org}
}

@inproceedings{Hong2025Aggregation,
  author       = {Xiaobin Hong and
                  Mingkai Lin and
                  Xiangkai Ma and
                  Wenzhong Li and
                  Sanglu Lu},
  title        = {Aggregation Mechanism Based Graph Heterogeneous Networks Distillation},
  booktitle    = {Proceedings of the Thirty-Fourth International Joint Conference on
                  Artificial Intelligence},
  pages        = {2901--2909},
  publisher    = {ijcai.org},
  year         = {2025},
  url          = {https://doi.org/10.24963/ijcai.2025/323},
  doi          = {10.24963/IJCAI.2025/323},
  bibsource    = {dblp computer science bibliography, https://dblp.org}
}

@inproceedings{Li2025Tackling,
  author       = {Moqi Li and
                  Xu Yang and
                  Cheng Deng},
  title        = {Tackling Long-Tailed Data Challenges in Spiking Neural Networks via
                  Heterogeneous Knowledge Distillation},
  booktitle    = {Proceedings of the Thirty-Fourth International Joint Conference on
                  Artificial Intelligence, {IJCAI} 2025, Montreal, Canada, August 16-22,
                  2025},
  pages        = {1404--1412},
  publisher    = {ijcai.org},
  year         = {2025},
  url          = {https://doi.org/10.24963/ijcai.2025/157},
  doi          = {10.24963/IJCAI.2025/157},
  bibsource    = {dblp computer science bibliography, https://dblp.org}
}

@inproceedings{Wei2023ELITE,
  author       = {Yuxiang Wei and
                  Yabo Zhang and
                  Zhilong Ji and
                  Jinfeng Bai and
                  Lei Zhang and
                  Wangmeng Zuo},
  title        = {{ELITE:} Encoding Visual Concepts into Textual Embeddings for Customized
                  Text-to-Image Generation},
  booktitle    = {{IEEE/CVF} International Conference on Computer Vision},
  pages        = {15897--15907},
  publisher    = {{IEEE}},
  year         = {2023},
  url          = {https://doi.org/10.1109/ICCV51070.2023.01461},
  doi          = {10.1109/ICCV51070.2023.01461},
  bibsource    = {dblp computer science bibliography, https://dblp.org}
}

\end{document}